\documentclass{article}
\usepackage{spconf,amsmath,graphicx}
\usepackage{times}
\usepackage{epsfig}
\usepackage{graphicx}
\usepackage{amsmath}
\usepackage{amssymb}
\usepackage{verbatim}
\usepackage{mathtools,xparse}
\usepackage{color, colortbl}
\usepackage{subcaption}

\definecolor{darkBlue}{rgb}{0.5,0.6,1}
\definecolor{LightCyan}{rgb}{0.5,0.8,1}
\definecolor{yellow}{rgb}{1,1,0.6}

\title{lightweight MONOCULAR DEPTH ESTIMATION model by JOINT END-TO-END FILTER PRUNING}
\name{Sara Elkerdawy \qquad Hong Zhang \qquad Nilanjan Ray}

\address{Department of Computing Science, University of Alberta}
\begin{document}
%
\maketitle
\begin{abstract}
   Convolutional neural networks (CNNs) have emerged as the state-of-the-art in multiple vision tasks including depth estimation. However, memory and computing power requirements remain as challenges to be tackled in these models. Monocular depth estimation has significant use in robotics and virtual reality that requires deployment on low-end devices. Training a small model from scratch results in a significant drop in accuracy and it does not benefit from pre-trained large models. Motivated by the literature of model pruning, we propose a lightweight monocular depth model obtained from a large trained model. This is achieved by removing the least important features with a novel joint end-to-end filter pruning. We propose to learn a binary mask for each filter to decide whether to drop the filter or not. These masks are trained jointly to exploit relations between filters at different layers as well as redundancy within the same layer. We show that we can achieve around 5x compression rate with small drop in accuracy on the KITTI driving dataset. We also show that masking can improve accuracy over the baseline with fewer parameters, even without enforcing compression loss.

\end{abstract}
\section{Introduction}
\label{sec:intro}

\begin{figure}[t]
\begin{center}
   \includegraphics[width=0.8\linewidth,height=2.5in]{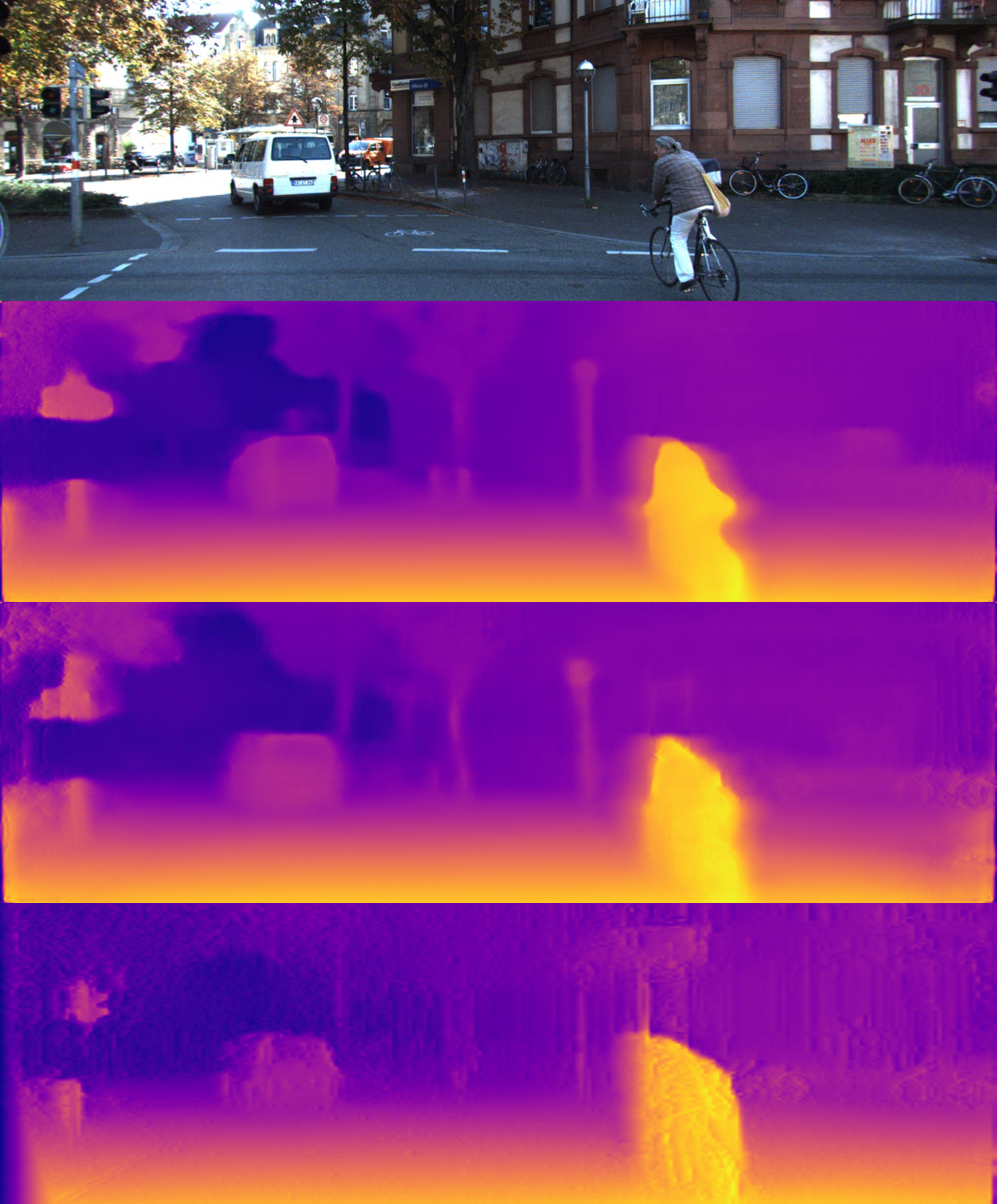}
\end{center}
   \caption{Example of our proposed pruning by masking. From top to bottom: LRC  \cite{godard2017unsupervised} (30.7M), ours (5.9M) and PyD-Net (1.9M) \cite{poggi2018towards}. Our network produces smooth similar prediction to LRC with 80.1\% reduction in number of parameters.}

\label{fig:intro}
\end{figure}

Real-time monocular depth estimation is needed in many visual tasks such as in robot simultaneous localization and mapping (SLAM), collision avoidance, and augmented reality. Monocular cameras are attractive as they are already deployed on many every-day systems such as phones, dash-cameras, and surveillance cameras. Although depth estimation from a single view is ill-posed due to geometric ambiguity, convolutional neural networks (CNN) achieved impressive results by solving the task as a learning problem. These methods rely on large and deep models and thus require high-end GPUs to run in real-time. This hinders deployment on many applications that require low-power consumption as well as real time response. While these issues can be tackled by designing small models, not only this requires expert knowledge and multiple manual tuning, but also these small models do not benefit from the trained large ones. As shown in \cite{hinton2012improving} and multiple work on distillation and pruning \cite{he2017channel,hinton2015distilling}, over-parameterization in training or guidance from large (teacher) models benefits the smaller model as a way of transfer learning. These methods achieve better accuracy compared to training the same small model from scratch. This gives rise to the idea of training large models and then applying pruning after training to get the smaller footprint, instead of training the small one from scratch.

In this paper, we propose to automatically generate thin models from trained larger ones in an end-to-end training to better scale up for large models as in depth estimation. We propose to train a binary mask for each convolutional filter that acts like a gate to drop the whole filter or not. In training, we encourage smaller models through inducing sparsity by minimizing the $\ell_1$-norm of the masks. To take into account the task loss as well, the masks are trained with both $\ell_1$ loss and the depth estimation loss. Closest to our proposed pruning method is \cite{he2017channel} in the masking aspect, however, we must emphasis that we learn the masks jointly on the whole network without prior knowledge on the compression rate per layer or computing layerwise sensitivity analysis as in \cite{li2016pruning}. In \cite{he2017channel}, the authors set the compression rate for each layer and adopt layer by layer pruning where each prune is followed by a finetune. Both of these points obstruct scalability for large datasets and models such as encoding-decoding architectures which are twice the size of classification models. All of these issues motivated us to train an end-to-end joint pruning method that can be adopted in large scale models and datasets suitable for depth estimation.

Our baseline deep models are trained based on LRC \cite{godard2017unsupervised} casting the problem as image reconstruction from stereo input pairs in an unsupervised setup. Figure \ref{fig:intro} shows an example output of our pruned model (5x size reduction), LRC and PyD-Net \cite{poggi2018towards}, a small-sized network. The pruned network produces smoother and more accurate depth prediction thanks to the pre-trained large model compared to similar small-sized models trained from scratch.

\begin{figure*}
\begin{center}
 \includegraphics[width=0.8\linewidth]{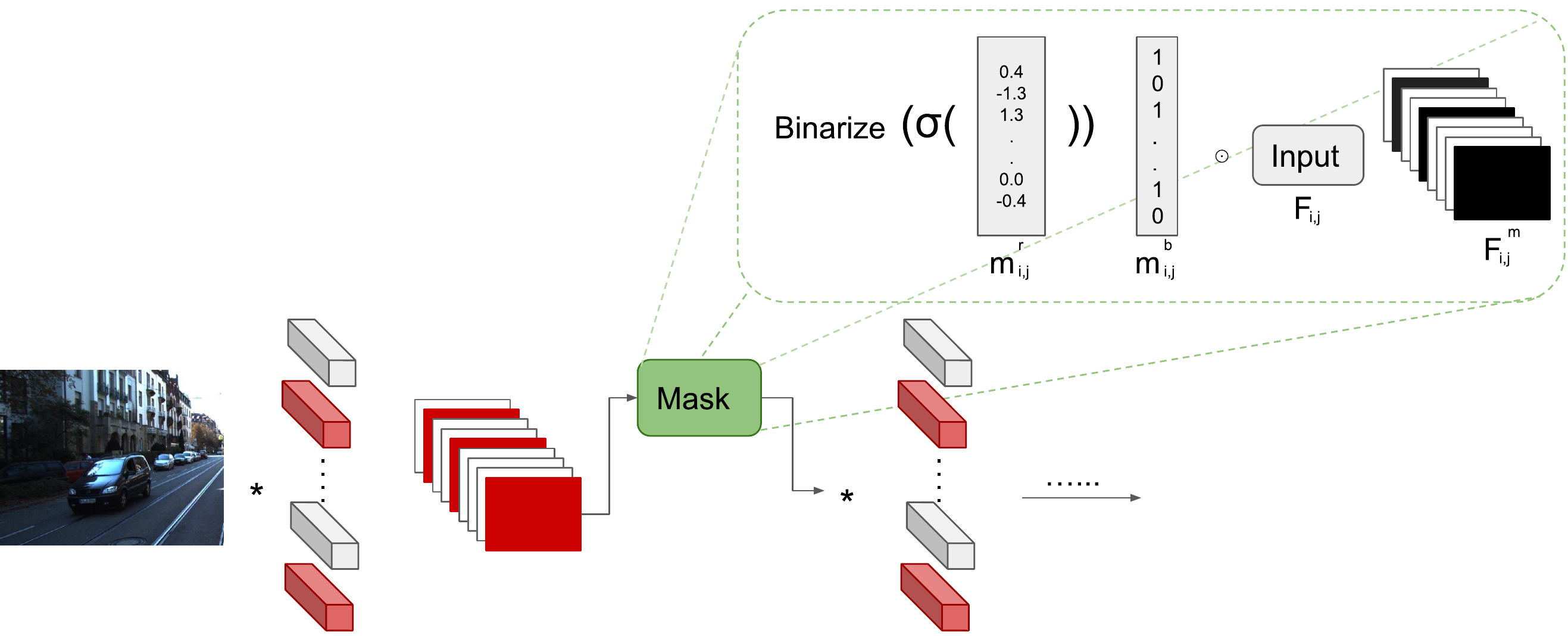}
\end{center}
   \caption{Architecture of the proposed joint end-to-end pruning. As in regular CNNs, image is passed through cascade of convolutional layers. A real-valued mask $m_{i,j}^r$ is learned through STE \cite{bengio2013estimating} from its corresponding binary $m_{i,j}^b$ estimation. The binary mask is multiplied by the input feature map $F_{i,j}$ to drop the corresponding filter contribution through forward and backward passes. The new masked feature maps $F_{i,:}^m$ (e.g with features zeroed out) are the new input for the next layer. We apply sigmoid function $\sigma$ on $m_{i,j}^r$ to limit the range of the real-values and simplify threshold selection in binarize function.}
\label{fig:short}
\end{figure*}
\section{Related Work}

In this section, we review literature on monocular depth estimation with supervised and unsupervised training.

\textbf{Supervised monocular depth estimation.} Methods in the supervised category directly estimate depth for each input pixel in the image given the ground-truth depth. Planar based approximation methods \cite{make3d,hoiem2005automatic} estimate local planes in the scene to predict 3D location and orientation for segmented patches. 
Eigen et al. \cite{eigen2015predicting} proposed a CNN model to directly infer the depth for each pixel in which lots of work followed this line of research \cite{cao2017estimating,ummenhofer2017demon}. Ummenhofer et al. \cite{ummenhofer2017demon} proposed a deep model to infer both ego-motion from sequence of frames to leverage motion in the depth estimation. Finally, Fu et al. \cite{fu2018deep} proposed to discretize depth and cast the problem as a multi-class classification with ordinal regression loss. However, for all these methods, the supervised paradigm requires a large amount of labeled data available to successfully learn a robust depth estimation. Acquiring such data is expensive and even with LiDAR data available, careful manual filtering of wrong projections \cite{geiger2012we} and removing the twist effect due to orbital nature of the sensor is still required.

\textbf{Unsupervised monocular depth estimation.} To avoid the issue of labeled data availability, recent work were proposed by posing the problem as image reconstruction. The ground-truth labels were replaced by different cues and losses from sequence of images \cite{zhou2017unsupervised} or view synthesize \cite{flynn2016deepstereo, xie2016deep3d}. Godard et al. \cite{godard2017unsupervised} samples the right image from learned disparity through differentiable bilinear module given stereo input. Zhou et al. \cite{zhou2017unsupervised} proposed to jointly predict the relative pose between unconstrained video sequences as well as computing the reconstruction loss between them. This has the advantage of removing the need of stereo pairs, but produces a less accurate final model due to moving objects. Finally closest to our focus, Poggi et al. \cite{poggi2018towards} proposed PyD-Net, a small sized CNN trained in similar setup as \cite{godard2017unsupervised} and allow for an affordable model on CPUs.

\section{Method overview}

In this section, we describe our end-to-end joint pruning method for monocular depth estimation. We base our solution on the unsupervised image reconstruction proposition by Godard et al. \cite{godard2017unsupervised}. 

The pipeline contains two main losses: 1) Task loss, and 2) sparsity loss. Task loss includes all losses with image reconstruction, disparity smoothness, and lr-consistency. The sparsity loss is applied on the masks with $\ell_1$-norm to encourage our model with fewer features.

\textbf{Task loss}. We train all the models with three weighted losses contributing to the final task loss as formulated in \cite{godard2017unsupervised}. 
\begin{equation}
\resizebox{.9\hsize}{!}{$L_{task}$ = $\alpha_{ap}$($L_{ap}^l$+$L_{ap}^r$) + $\alpha_{ds}$($L_{ds}^l$+$L_{ds}^r$) + $\alpha_{lr}$($L_{lr}^l$+$L_{lr}^r$)}
\label{eq:task}
\end{equation}
Each loss term is calculated for both left and right images in the stereo input pair. The first term $L_{ap}$ calculates the reconstruction loss between the original image and the warped image using SSIM \cite{wang2004image} and $L1$ difference. The second term $L_{ds}$ encourages disparity discontinuities only at gradient of image $\delta I$. Finally, the left-right consistency term $L_{lr}$ enforces coherence between predicted left disparity $d^l$ and predicted right disparity $d^r$.

\textbf{Mask sparsity loss}. This loss term controls the model size. Before diving into the sparsity loss, we explain the masking formulation. First, we initialize real-valued mask $m_i^r \in \mathbb{R}^{n_i}$ for each layer $i$ with $n_i$ filters. A binary function is then applied on the real-valued masks to get $m_i^b$ based on a threshold $t$ (e.g $m_{i,j}^r$ $\geqslant t$ outputs 1 and 0 otherwise). Finally, $\ell_1$ is applied on all the masks to form a sparsity loss term:
\begin{equation} \label{eq:sparse}
L_{mask} = \frac{\sum_{i}^{N}{\parallel{m_i^b}\parallel_1}}{\sum_{i}^{N}{n_i}}
\end{equation}
where $N$ is the total number of layers in the network. Looking carefully at (\ref{eq:sparse}), as $m_i^b$ vectors are binary, the loss term is calculating the ratio between the total number of filters in the new model and the original large model. Minimizing this loss is equivalent to maximizing the compression rate. Our total loss is then given by
\begin{equation}
    L_{total} = L_{task} + L_{mask}
\end{equation}

\textbf{Forward pass.} Let $F_{i,j}$ be the $j$th feature map of the $i$th layer, the new feature map $F_{i,j}^m$ and binary mask $m_{i,j}^b$ are thus given by:
\begin{equation} \label{eq:mr}
\begin{gathered}
   m_{i,j}^b = Binarize(Sigmoid(m_{i,j}^r),0.5)\\
   F_{i,j,h,w}^m = F_{i,j,h,w} \odot m_{i,j}^b
\end{gathered} 
\end{equation}

We apply a sigmoid function on our real-valued masks to transfer the input into [0,1] range before passing though binarization. This simplifies the selection of threshold $t$ as a sensible choice would be 0.5. Finally, $F_{i,j}^m$ is passed as the new (${i+1}$)th layer's input which corresponds to either $F_{i,j}$ or 0. Zeroing out a feature map $F_{i,j}$ simulates dropping the corresponding filter $f_{i,j}$. Figure \ref{fig:short} shows the masking block embedded within the network summarizing the forward pass.

\textbf{Backward pass.} 
In the backward pass, we update the convolutional kernels and $m_{i,j}^r$ for each layer. As $Binarize$ is a conditional non-differentiable function, we backpropagate the gradients to $m_{i,j}^r$ utilizing the straight-through-estimator (STE) proposed in \cite{bengio2013estimating}. They showed that we can approximate the gradient of a real valued weight with the gradient of its discretization. Even though gradients calculated through such a function (e.g $Binarize$) are noisy, they serve as regularizers and are acceptable approximation of the true gradients to the real-valued masks. Using STE and from \ref{eq:mr}, we have the gradients as:
\begin{equation}
\begin{gathered}
   \delta{m_{i,j}^b} \triangleq \frac{\partial L_{total}}{\partial m_{i,j}^b} = \sum_h^H \sum_w^W \delta{F_{i,j,h,w}^m} \cdot F_{i,j,h,w} \\
   \resizebox{0.9\columnwidth}{!}{$\delta{m_{i,j}^r}$ = $\delta{m_{i,j}^b} \cdot Sigmoid(m_{i,j}^r) \cdot (1-Sigmoid(m_{i,j}^r))$}
\end{gathered}
\end{equation}
The double sum stems from the fact that $m_{i,j}^b$ is shared among all spatial locations in $F_{i,j}$.

\section{Experiments and analysis}

\begin{figure*}[t]
\centering
     \includegraphics[width=0.3\textwidth,height=1cm]{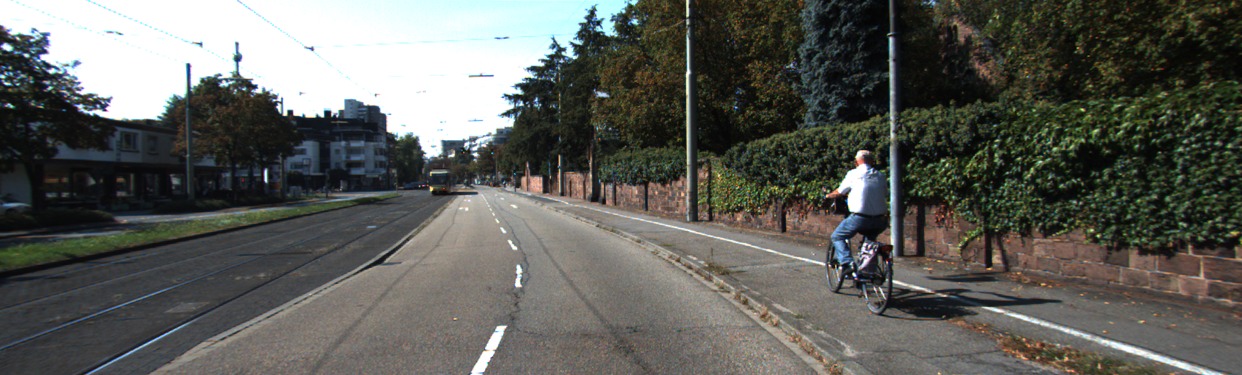}
    \includegraphics[width=0.3\textwidth,height=1cm]{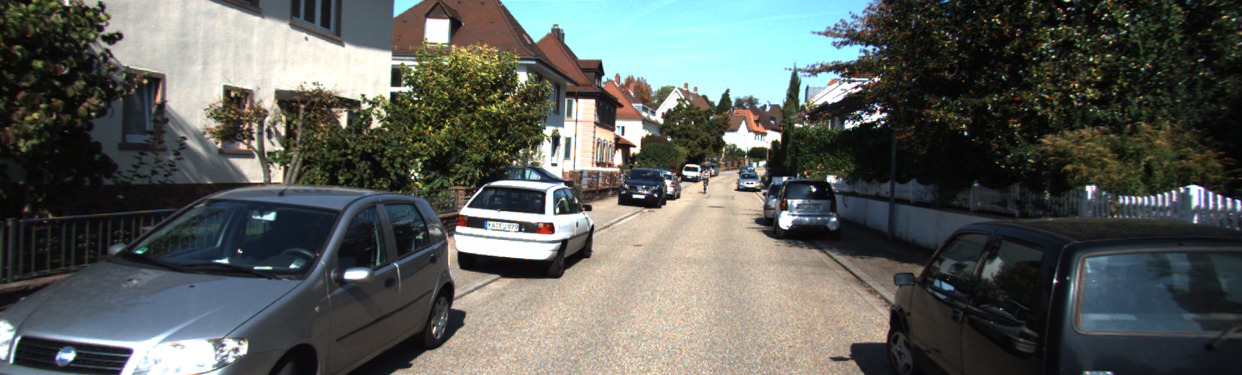}
    \includegraphics[width=0.3\textwidth,height=1cm]{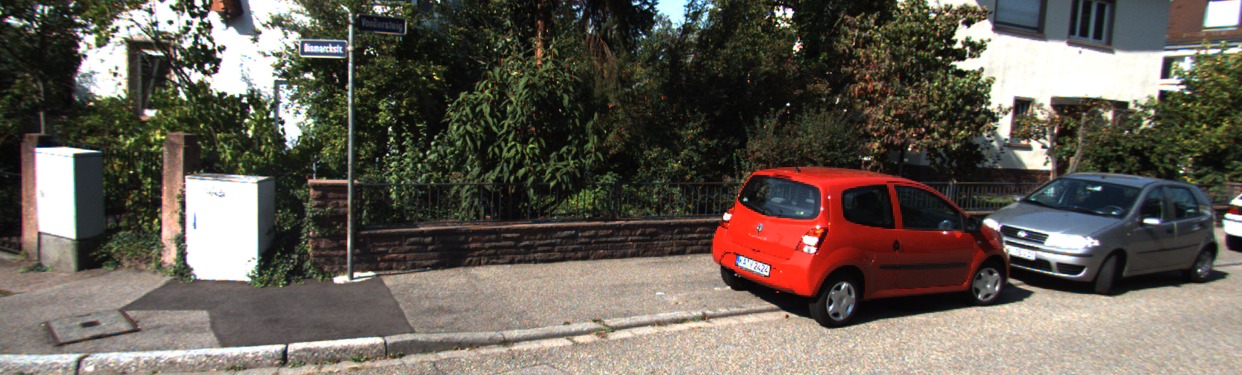}
    
    \includegraphics[width=0.3\textwidth,height=1cm]{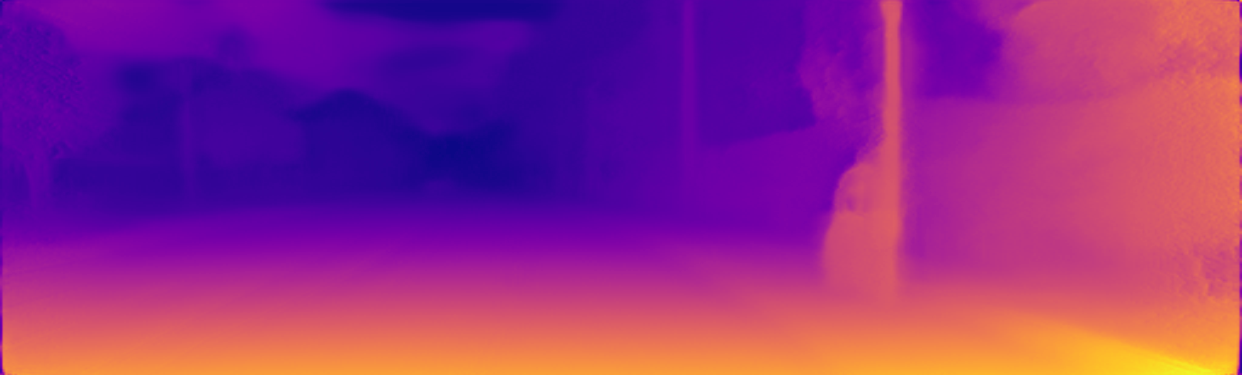}
    \includegraphics[width=0.3\textwidth,height=1cm]{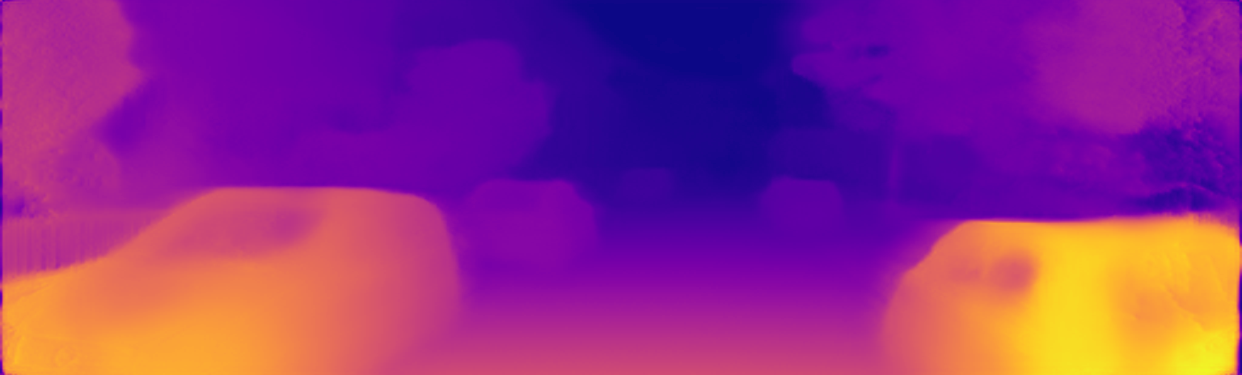}
    \includegraphics[width=0.3\textwidth,height=1cm]{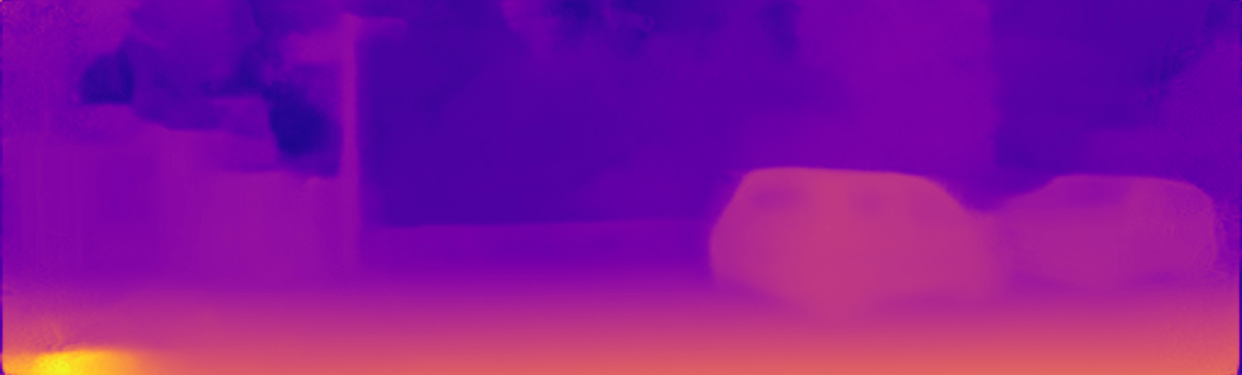}
    
    \includegraphics[width=0.3\textwidth,height=1cm]{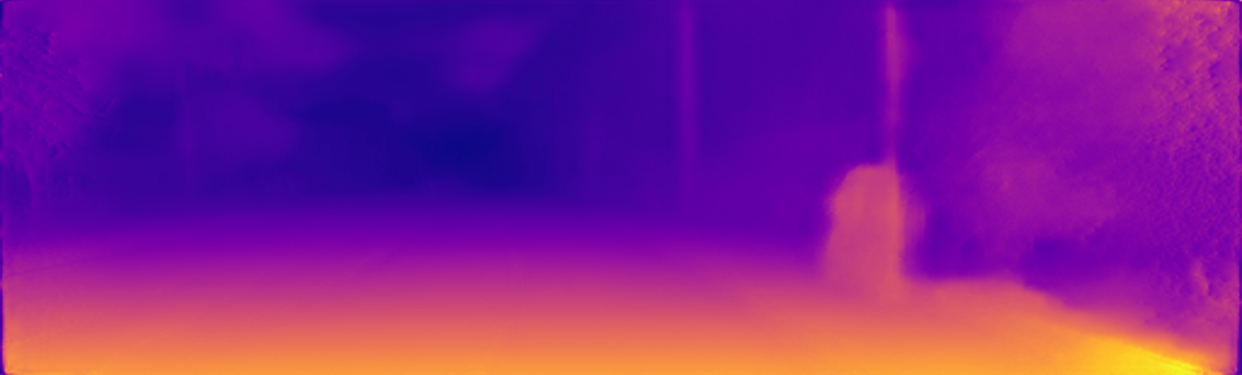}
    \includegraphics[width=0.3\textwidth,height=1cm]{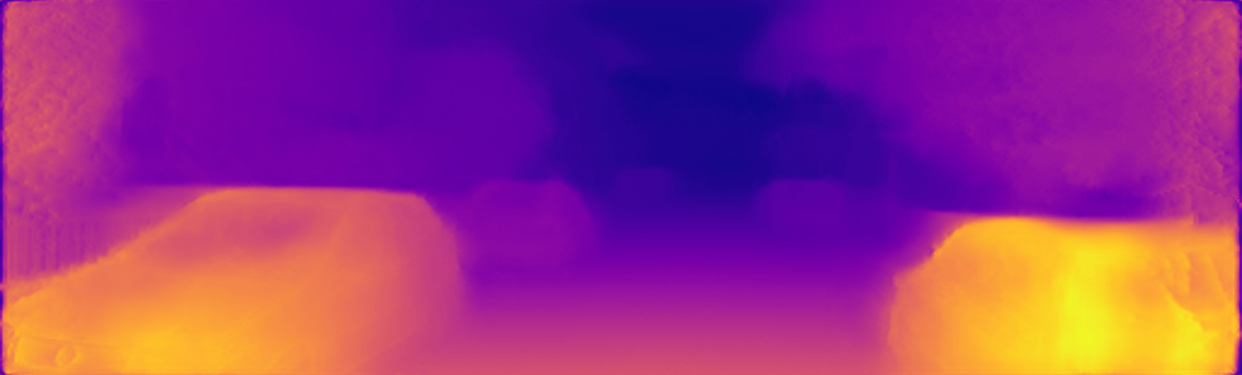}
    \includegraphics[width=0.3\textwidth,height=1cm]{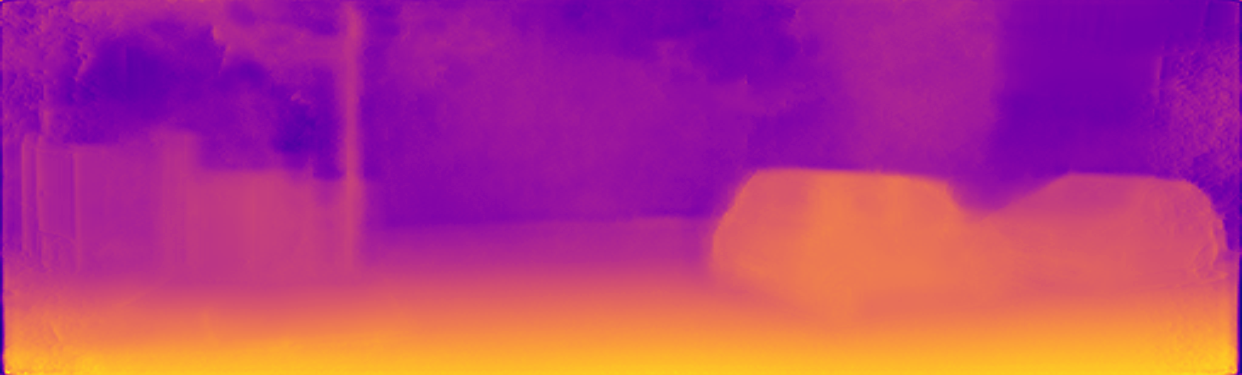}

    \includegraphics[width=0.3\textwidth,height=1cm]{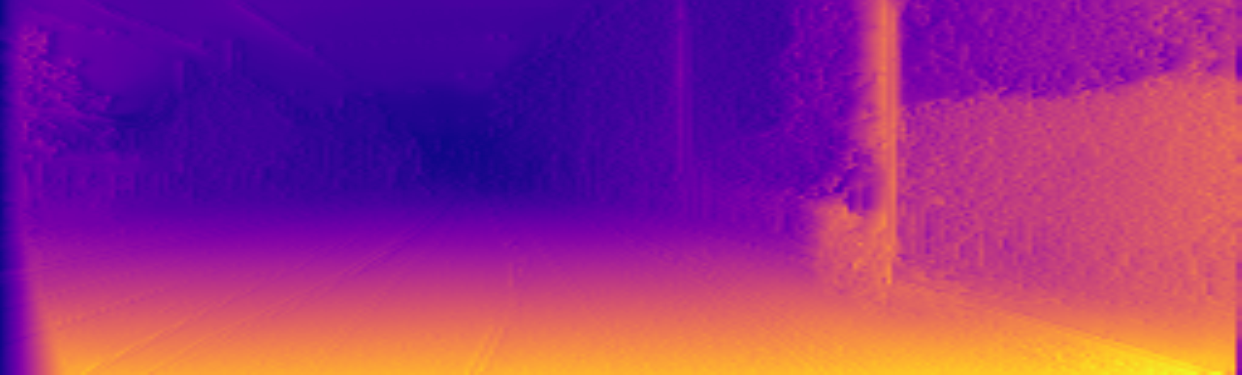}
    \includegraphics[width=0.3\textwidth,height=1cm]{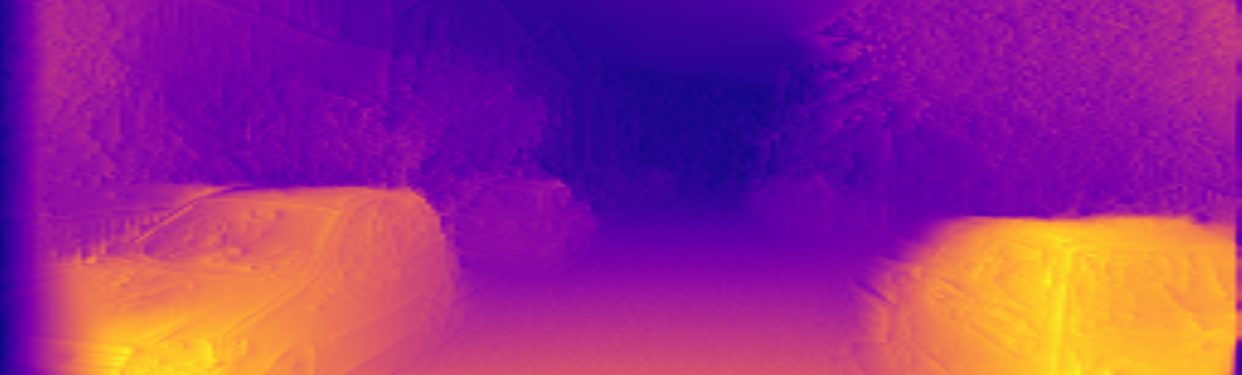}
    \includegraphics[width=0.3\textwidth,height=1cm]{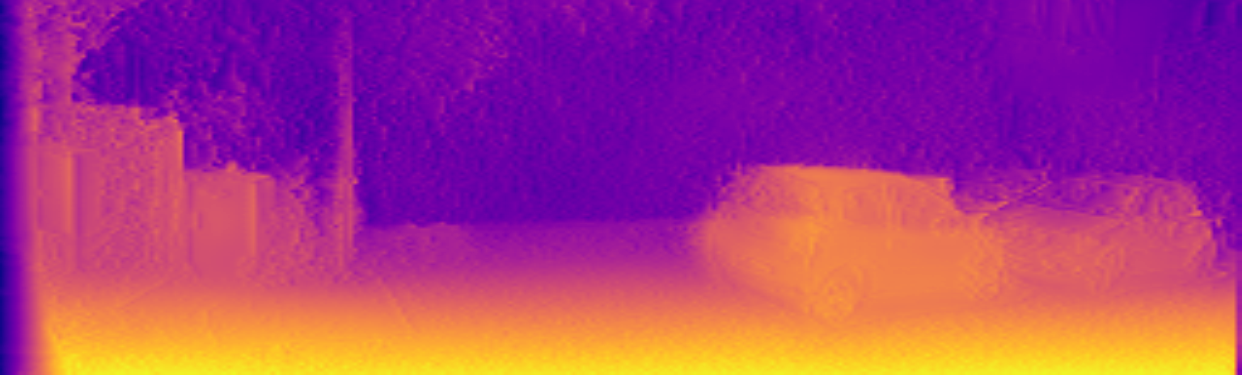}
    
    \caption{Depth predictions on KITTI Eigen compared with LRC \cite{godard2017unsupervised} 31.6M, ours VGG+$L_{total}$ 5.9M, PyD-Net 1.9M \cite{poggi2018towards} from top to bottom. Our pruned model produces good quality smooth output compared to PyD-Net but still with small accuracy drop (e.g pole in first column). Small models better regularize scenes with fewer data in the training (e.g a turn in third column)}
    \label{fig:depth}

\end{figure*}

\subsection{Experiments setup}
 We train all networks with 50 epochs, batch size 8, and Adam optimizer \cite{adam} with the default parameters $\beta_1 =$0.9 and $\beta_2 =$0.999. We set a learning rate scheduler that is initially set with $\lambda = 10^{-4}$ and is halved each 10 epochs after the first 30 epochs. Binary masks $m_{i,j}^b$ are initialized with 1 to keep all the filters. Finally, we perform data augmentation on the fly by randomly flipping the input images horizontally and applying image transformation on the color spectrum as in \cite{godard2017unsupervised}.

We evaluate our pipeline with state-of-the-art large deep models and small proposed models for monocular depth estimation. The evaluations are done on KITTI dataset \cite{geiger2012we} on both Eigen and KITTI splits. We also make use of Cityscapes dataset \cite{cordts2016cityscapes} which is a large collected dataset containing 22,973 training stereo pairs. However, the provided depth images are generated using SGM \cite{hirschmuller2011semi}, so not suitable for evaluation. We use it for pre-training and then fine-tuning on KITTI as previous work.

\begin{table}[htb]
  \centering
  \resizebox{\columnwidth}{!}{\begin{tabular}{|c|c|c|c|}
  \multicolumn{1}{c}{} & \multicolumn{1}{c}{\cellcolor{yellow}Ours} & \multicolumn{1}{c}{\cellcolor{LightCyan}Lower is better} & \multicolumn{1}{c}{}\\
  \hline
  Method  & Dataset & \cellcolor{LightCyan}D1 all & Params\\
  \hline
LRC + Deep3D \cite{godard2017unsupervised}& K & 59.64 & 31.6M \\

LRC + VGG \cite{godard2017unsupervised}& K & 30.272 & 31.6M \\

 \rowcolor{yellow}VGG + $L_{total}$ & K & 32.183 $\uparrow$ 1.9 & 7.5M $\downarrow$ 76.1\% \\

LRC + Resnet50 \cite{godard2017unsupervised}& K & 28.459 & 58.4M \\

PyD-Net \cite{poggi2018towards}  & K & 38.478 & 1.9M \\
\hline
LRC + VGG \cite{godard2017unsupervised}& CS + K & 25.523 & 31.6M \\

\rowcolor{yellow}VGG + $L_{task}$ & CS + K & 24.939 $\downarrow$ 1.33 & 30.2M $\downarrow$ 4.2\%\\

\rowcolor{yellow}VGG + $L_{total}$ & CS + K & 26.861 $\uparrow$ 1.3 & 4.3M $\downarrow$ 86.1\%\\

LRC + VGG pp* \cite{godard2017unsupervised} & CS + K & 25.077 & 31.6M 2x forward\\

LRC + Resnet50 \cite{godard2017unsupervised} &  CS + K & 24.504 & 58.4M \\
\hline
  \end{tabular}}
 
  \caption{Comparison of different models on KITTI 2015 stereo split. For training, K is \cite{geiger2012we} and CS is Cityscapes \cite{cordts2016cityscapes}. Our models prune more than 76\% of the original model with maximum 1.9\% drop in accuracy. *pp is post-processing done by \cite{godard2017unsupervised} but requires two forward pass. Suffix $L_x$ in our method indicates the training loss used.}
  \label{tab:KITTI}
\end{table}

\begin{table}[htb]
  \centering
  \resizebox{\columnwidth}{!}{\begin{tabular}{|c|c|c|c|c|c|c|c|c|c|}
  \multicolumn{1}{c}{} & \multicolumn{1}{c}{\cellcolor{yellow}Ours} & \multicolumn{2}{c}{\cellcolor{LightCyan}Lower is better} & \multicolumn{2}{c}{\cellcolor{darkBlue}Higher is better} \\
  \hline
  Method   & Dataset & \cellcolor{LightCyan}Abs Rel & \cellcolor{LightCyan}RMS & \cellcolor{darkBlue}$\delta < 1.25$ & Params\\
  \hline
  Eigen et al. \cite{eigen2015predicting}  & K & 0.203 & 6.307 & 0.702 & 54.2M \\
  Liu et al. \cite{liu2016learning} & K & 0.201 & 6.471 & 0.680 & 40.0M \\
  Zhou et al. \cite{zhou2017unsupervised} & K & 0.208 & 6.856 & 0.678 & 34.2M \\
  LRC + VGG \cite{godard2017unsupervised} & K & 0.148 & 5.927 & 0.803 & 31.6M \\
  \rowcolor{yellow}VGG + $L_{total}$ & K & 0.1356 & 5.891 & 0.827 & 5.7M $\downarrow$ 81.8\%\\
  PyD-Net & K & 0.163 & 6.253 & 0.759 & 1.9M \\
  \hline
  LRC + VGG \cite{godard2017unsupervised} & CS+K & 0.124 & 5.311 & 0.847 & 31.6\\
  \rowcolor{yellow}VGG + $L_{task}$ & CS+K & 0.124 & 5.280 $\downarrow$ 0.03& 0.848 & 30.8M $\downarrow$ 2\%\\
  \rowcolor{yellow}VGG + $L_{total}$ & CS+K & 0.1452& 5.835 $\uparrow$ 0.524& 0.815 & 5.9M $\downarrow$ 81.1\% \\
  LRC + ResNet50 pp* \cite{godard2017unsupervised} & CS+K & 0.114 & 4.935 & 0.861 & 58.4M\\
  PyD-Net \cite{poggi2018towards} & CS+K & 0.148 & 5.929 & 0.800 & 1.9M\\
\hline
\end{tabular}}
 
  \caption{Comparison on Eigen split. For training, K is \cite{geiger2012we} and CS is Cityscapes \cite{cordts2016cityscapes}. Our models compress more than 81.1\% the original model with small drop in accuracy. *pp post-processing done by \cite{godard2017unsupervised} but requires two forward passes.}
  \label{tab:Eigen}
\end{table}

\subsection{Evaluation}
Results are compared using depth metrics from \cite{eigen2015predicting} but some are discarded due to limited space.\\
\textbf{KITTI split.} Table \ref{tab:KITTI} shows the comparison on KITTI split with different deep models. As shown, our end-to-end pruning method achieved around 5x reduction in the number of parameters with small drop in accuracy. As the compression rate is optimized within the network, obtaining exact number of parameters for fair comparison with same sized models is hard. Although PyD-Net has fewer parameters, our method was able to generate an on par small sized model without manually engineering the model footprint. Interestingly, training the network with masks without enforcing compressing by optimizing only $L_{task}$ as described in Eq.\ref{eq:task} provided a form of regularization to the network and improved the accuracy of the original trained network. This is similar to dropout \cite{srivastava2014dropout} as dropping features in the training breaks co-adaptation between the features. VGG + $L_{task}$ outperforms VGG LRC with post-processing which requires double the time for two forward passes and achieves similar accuracy to Resnet50 LRC with 2x less parameters.

\textbf{Eigen split.} Table \ref{tab:Eigen} shows evaluation on Eigen split with the other methods reporting on Eigen for completeness even with the noisy quality of the LiDAR ground truth. Our smaller models achieve better accuracy than the supervised methods \cite{eigen2015predicting,liu2016learning} and unsupervised method \cite{zhou2017unsupervised}. Interestingly, the gap in accuracy (e.g 5th column) between our pruned model and PyD-Net differs based on the training data. As small models require large amount of data (e.g CS+K) to achieve good results, our method on the other hand benefits from the pre-trained large model even when trained with KITTI dataset only. This shows the benefit from pruning rather than training from scratch specially with limited training data.

\textbf{Qualitative results}
Figure \ref{fig:depth} shows some qualitative comparison to LRC \cite{godard2017unsupervised} and PyDNet \cite{poggi2018towards}. Although our pruned model is 5x smaller than LRC, they still produce similar good quality smooth output. Our model benefits from the pre-trained VGG model to produce smooth output and not as noisy as the case with similar small sized model PyDNet. It is worth noting that small models (ours and PyDNet) better regularize scenes with fewer data in the training unlike LRC as shown in third column. However, the pruned model shows accuracy drop with small objects (e.g poles).

\subsection{Conclusion}

We proposed a lightweight model for monocular depth estimation motivated by pruning literature. Our joint end-to-end pruning is scalable for deep models adopted in depth estimation. We learn binary masks within the network to drop filters jointly without predefining layerwise compression rates. We showed how pruning benefits small model training compared to training from scratch specially with limited data.

\bibliographystyle{IEEEbib}
\bibliography{strings}

\begin{thebibliography}{10}

\bibitem{godard2017unsupervised}
Cl{\'e}ment Godard, Oisin Mac~Aodha, and Gabriel~J Brostow,
\newblock ``Unsupervised monocular depth estimation with left-right
  consistency,''
\newblock in {\em CVPR}, 2017, vol.~2, p.~7.

\bibitem{poggi2018towards}
Matteo Poggi, Filippo Aleotti, Fabio Tosi, and Stefano Mattoccia,
\newblock ``Towards real-time unsupervised monocular depth estimation on cpu,''
\newblock {\em arXiv preprint arXiv:1806.11430}, 2018.

\bibitem{hinton2012improving}
Geoffrey~E Hinton, Nitish Srivastava, Alex Krizhevsky, Ilya Sutskever, and
  Ruslan~R Salakhutdinov,
\newblock ``Improving neural networks by preventing co-adaptation of feature
  detectors,''
\newblock {\em arXiv preprint arXiv:1207.0580}, 2012.

\bibitem{he2017channel}
Yihui He, Xiangyu Zhang, and Jian Sun,
\newblock ``Channel pruning for accelerating very deep neural networks,''
\newblock in {\em International Conference on Computer Vision (ICCV)}, 2017,
  vol.~2.

\bibitem{hinton2015distilling}
Geoffrey Hinton, Oriol Vinyals, and Jeff Dean,
\newblock ``Distilling the knowledge in a neural network,''
\newblock {\em arXiv preprint arXiv:1503.02531}, 2015.

\bibitem{li2016pruning}
Hao Li, Asim Kadav, Igor Durdanovic, Hanan Samet, and Hans~Peter Graf,
\newblock ``Pruning filters for efficient convnets,''
\newblock {\em arXiv preprint arXiv:1608.08710}, 2016.

\bibitem{bengio2013estimating}
Yoshua Bengio, Nicholas L{\'e}onard, and Aaron Courville,
\newblock ``Estimating or propagating gradients through stochastic neurons for
  conditional computation,''
\newblock {\em arXiv preprint arXiv:1308.3432}, 2013.

\bibitem{make3d}
Ashutosh Saxena, Min Sun, and Andrew~Y Ng,
\newblock ``Make3d: Learning 3d scene structure from a single still image,''
\newblock {\em IEEE transactions on pattern analysis and machine intelligence},
  vol. 31, no. 5, pp. 824--840, 2009.

\bibitem{hoiem2005automatic}
Derek Hoiem, Alexei~A Efros, and Martial Hebert,
\newblock ``Automatic photo pop-up,''
\newblock in {\em ACM transactions on graphics (TOG)}. ACM, 2005, vol.~24, pp.
  577--584.

\bibitem{eigen2015predicting}
David Eigen and Rob Fergus,
\newblock ``Predicting depth, surface normals and semantic labels with a common
  multi-scale convolutional architecture,''
\newblock in {\em Proceedings of the IEEE International Conference on Computer
  Vision}, 2015, pp. 2650--2658.

\bibitem{cao2017estimating}
Yuanzhouhan Cao, Zifeng Wu, and Chunhua Shen,
\newblock ``Estimating depth from monocular images as classification using deep
  fully convolutional residual networks,''
\newblock {\em IEEE Transactions on Circuits and Systems for Video Technology},
  2017.

\bibitem{ummenhofer2017demon}
Benjamin Ummenhofer, Huizhong Zhou, Jonas Uhrig, Nikolaus Mayer, Eddy Ilg,
  Alexey Dosovitskiy, and Thomas Brox,
\newblock ``Demon: Depth and motion network for learning monocular stereo,''
\newblock in {\em IEEE Conference on computer vision and pattern recognition
  (CVPR)}, 2017, vol.~5, p.~6.

\bibitem{fu2018deep}
Huan Fu, Mingming Gong, Chaohui Wang, Kayhan Batmanghelich, and Dacheng Tao,
\newblock ``Deep ordinal regression network for monocular depth estimation,''
\newblock in {\em Proceedings of the IEEE Conference on Computer Vision and
  Pattern Recognition}, 2018, pp. 2002--2011.

\bibitem{geiger2012we}
Andreas Geiger, Philip Lenz, and Raquel Urtasun,
\newblock ``Are we ready for autonomous driving? the kitti vision benchmark
  suite,''
\newblock in {\em Computer Vision and Pattern Recognition (CVPR), 2012 IEEE
  Conference on}. IEEE, 2012, pp. 3354--3361.

\bibitem{zhou2017unsupervised}
Tinghui Zhou, Matthew Brown, Noah Snavely, and David~G Lowe,
\newblock ``Unsupervised learning of depth and ego-motion from video,''
\newblock in {\em CVPR}, 2017, vol.~2, p.~7.

\bibitem{flynn2016deepstereo}
John Flynn, Ivan Neulander, James Philbin, and Noah Snavely,
\newblock ``Deepstereo: Learning to predict new views from the world's
  imagery,''
\newblock in {\em Proceedings of the IEEE Conference on Computer Vision and
  Pattern Recognition}, 2016, pp. 5515--5524.

\bibitem{xie2016deep3d}
Junyuan Xie, Ross Girshick, and Ali Farhadi,
\newblock ``Deep3d: Fully automatic 2d-to-3d video conversion with deep
  convolutional neural networks,''
\newblock in {\em European Conference on Computer Vision}. Springer, 2016, pp.
  842--857.

\bibitem{wang2004image}
Zhou Wang, Alan~C Bovik, Hamid~R Sheikh, and Eero~P Simoncelli,
\newblock ``Image quality assessment: from error visibility to structural
  similarity,''
\newblock {\em IEEE transactions on image processing}, vol. 13, no. 4, pp.
  600--612, 2004.

\bibitem{adam}
Diederik~P. Kingma and Jimmy Ba,
\newblock ``Adam: {A} method for stochastic optimization,''
\newblock {\em CoRR}, vol. abs/1412.6980, 2014.

\bibitem{cordts2016cityscapes}
Marius Cordts, Mohamed Omran, Sebastian Ramos, Timo Rehfeld, Markus Enzweiler,
  Rodrigo Benenson, Uwe Franke, Stefan Roth, and Bernt Schiele,
\newblock ``The cityscapes dataset for semantic urban scene understanding,''
\newblock in {\em Proceedings of the IEEE conference on computer vision and
  pattern recognition}, 2016, pp. 3213--3223.

\bibitem{hirschmuller2011semi}
Heiko Hirschm{\"u}ller,
\newblock ``Semi-global matching-motivation, developments and applications,''
\newblock in {\em Photogrammetric week}, 2011, vol.~11, pp. 173--184.

\bibitem{liu2016learning}
Fayao Liu, Chunhua Shen, Guosheng Lin, and Ian~D Reid,
\newblock ``Learning depth from single monocular images using deep
  convolutional neural fields.,''
\newblock {\em IEEE Trans. Pattern Anal. Mach. Intell.}, vol. 38, no. 10, pp.
  2024--2039, 2016.

\bibitem{srivastava2014dropout}
Nitish Srivastava, Geoffrey Hinton, Alex Krizhevsky, Ilya Sutskever, and Ruslan
  Salakhutdinov,
\newblock ``Dropout: a simple way to prevent neural networks from
  overfitting,''
\newblock {\em The Journal of Machine Learning Research}, vol. 15, no. 1, pp.
  1929--1958, 2014.

\end{thebibliography}

\end{document}


{\centering \LARGE Supplementary Materials \par}
	
\section{Evaluation} Full depth metrics tables for KITTI and Eigen splits are shown in Table \ref{tab:KITTI} and Table \ref{tab:Eigen}
\begin{table*}[ht]
  \centering
  \resizebox{\textwidth}{!}{\begin{tabular}{|c|c|c|c|c|c|c|c|c|c|c|}
  \multicolumn{2}{c}{\cellcolor{yellow}Ours} & \multicolumn{5}{c}{\cellcolor{LightCyan}Lower is better} & \multicolumn{3}{c}{\cellcolor{darkBlue}Higher is better} & \multicolumn{1}{c}{}\\
  \hline
  Method  & Dataset & \cellcolor{LightCyan}Abs Rel & \cellcolor{LightCyan}Sq Rel & \cellcolor{LightCyan}RMS &  \cellcolor{LightCyan}RMS_{log} & \cellcolor{LightCyan}D1 all & \cellcolor{darkBlue}\delta < 1.25 & \cellcolor{darkBlue}\delta < 1.25^2 & \cellcolor{darkBlue}\delta < 1.25^3 & Params\\
  \hline
LRC + Deep3D \cite{godard2017unsupervised}& K & 0.151 & 1.312 & 6.344 & 0.239 & 59.64 & 0.781 & 0.931 & 0.976 & 31.6M \\

LRC + VGG \cite{godard2017unsupervised}& K & 0.124 & 1.388 & 6.125 & 0.217 & 30.272 & 0.841 & 0.936 & 0.975 & 31.6M \\

 \rowcolor{yellow}VGG + L_{total} & K &  0.1313 & 1.5296 & 6.488 & 0.230 & 32.183 \uparrow 1.9 & 0.826 & 0.927 & 0.970 & 7.5M \downarrow 76.1\% \\

LRC + Resnet50 \cite{godard2017unsupervised}& K & 0.1139 & 1.2661 & 5.784 & 0.204 & 28.459 & 0.853 & 0.947 & 0.979 & 58.4M \\

PyD-Net \cite{poggi2018towards}  & K & 0.1393 &  1.4720 & 6.570 & 0.240 & 38.478 & 0.805  & 0.919 & 0.967 & 1.9M \\
\hline
LRC + VGG \cite{godard2017unsupervised}& CS + K & 0.104 & 1.070 & 5.417 & 0.188 & 25.523 & 0.875 & 0.956 & 0.983 & 31.6M \\

\rowcolor{yellow}VGG + L_{task} & CS + K & 0.1026 & 1.0471 & 5.366 & 0.187 & 24.939 \downarrow 1.33& 0.874 & 0.957 & 0.983    & 30.2M \downarrow 4.2\%\\

\rowcolor{yellow}VGG + L_{total} & CS + K & 0.1101 & 1.1562 & 5.688 & 0.196 & 26.861 \uparrow 1.3 & 0.865 & 0.950 & 0.980 & 4.3M \downarrow 86.1\%\\

LRC + VGG pp* \cite{godard2017unsupervised} & CS + K & 0.100 & 0.934 & 5.141 & 0.178 & 25.077 & 0.878 & 0.961 & 0.986 & 31.6M 2x forward\\

LRC + Resnet50 \cite{godard2017unsupervised} &  CS + K & 0.1009 & 1.0315 & 5.360 &  0.184 & 24.504 & 0.878 & 0.959 & 0.983 & 58.4M \\
\hline
  \end{tabular}}
 
  \caption{Comparison of different models on KITTI 2015 stereo split. In dataset, K is \cite{geiger2012we} and CS is Cityscapes \cite{cordts2016cityscapes}. Our models prune more than 76\% of the original model with maximum 1.9\% drop in accuracy.  D1-all score represents the percentage of pixels having a disparity error larger than 3. *pp is post-processing done by \cite{godard2017unsupervised} but requires two forward pass. Suffix $L_x$ in our method indicates the training loss used.}
  \label{tab:KITTI}
\end{table*}

\begin{table*}[ht]
  \centering
  \resizebox{\textwidth}{!}{\begin{tabular}{|c|c|c|c|c|c|c|c|c|c|c|}
  \multicolumn{1}{c}{} & \multicolumn{2}{c}{\cellcolor{yellow}Ours} & \multicolumn{4}{c}{\cellcolor{LightCyan}Lower is better} & \multicolumn{3}{c}{\cellcolor{darkBlue}Higher is better} & \multicolumn{1}{c}{}\\
  \hline
  Method  & Supervised & Dataset & \cellcolor{LightCyan}Abs Rel & \cellcolor{LightCyan}Sq Rel & \cellcolor{LightCyan}RMS &  \cellcolor{LightCyan}RMS_{log} & \cellcolor{darkBlue}\delta < 1.25 & \cellcolor{darkBlue}\delta < 1.25^2 & \cellcolor{darkBlue}\delta < 1.25^3 & Params\\
  \hline
  Eigen et al. \cite{eigen2015predicting} & Yes & K & 0.203 & 1.548 & 6.307 & 0.282 & 0.702 & 0.890 & 0.958 & 54.2M \\
  Liu et al. \cite{liu2016learning}& Yes & K & 0.201 & 1.584 & 6.471 & 0.273 & 0.680 & 0.898 & 0.967 & 40.0M \\
  Zhou et al. \cite{zhou2017unsupervised}& No & K & 0.208 & 1.768 & 6.856 & 0.283 & 0.678 & 0.885 & 0.957 & 34.2M \\
  LRC + VGG \cite{godard2017unsupervised}& No & K & 0.148 & 1.344 & 5.927 & 0.247 & 0.803 & 0.922 & 0.964 & 31.6M \\
  \rowcolor{yellow}VGG + L_{total} & No & K & 0.1356 & 1.3625 & 5.891 & 0.236 & 0.827 & 0.927 & 0.965 & 5.7M \downarrow 81.8\%\\
  PyD-Net & No & K & 0.163 & 1.399 & 6.253 & 0.262 & 0.759 & 0.911 & 0.961 & 1.9M \\
  \hline
  Zhou et al. \cite{zhou2017unsupervised}& No & CS+K & 0.198 & 1.836 & 6.565 & 0.275 & 0.718 & 0.901 & 0.960 & 34.2M\\
  LRC + VGG \cite{godard2017unsupervised}& No & CS+K & 0.124 & 1.076 & 5.311 & 0.219 & 0.847 & 0.942 & 0.973 & 31.6\\
  \rowcolor{yellow}VGG + L_{task} & No & CS+K & 0.124 --- & 1.0775 & 5.280 \downarrow 0.03& 0.219 & 0.848 & 0.942 & 0.973 & 30.8M \downarrow 2\%\\
  \rowcolor{yellow}VGG + L_{total} & No & CS+K & 0.1452 \uparrow 0.02& 1.4024 & 5.835 \uparrow 0.524& 0.239 & 0.815 & 0.927 & 0.967 & 5.9M \downarrow 81.1\% \\
  LRC + ResNet50 pp* \cite{godard2017unsupervised} & No & CS+K & 0.114 & 0.898 & 4.935 & 0.206 & 0.861 & 0.949 & 0.976 & 58.4M 2x forward\\
  PyD-Net \cite{poggi2018towards} & No & CS+K & 0.148 & 1.316 & 5.929 & 0.244 & 0.800 & 0.925 & 0.967 & 1.9M\\
\hline
\end{tabular}}
 
  \caption{Comparison on Eigen split. In dataset, K indicates training on \cite{geiger2012we} and CS indicates Cityscapes \cite{cordts2016cityscapes}. Our models compress more than 80\% the original model with small drop in accuracy. *pp post-processing done by \cite{godard2017unsupervised} but requires two forward passes.}
  \label{tab:Eigen}
\end{table*}

\begin{figure*}[t]
\centering
\begin{subfigure}{.45\textwidth}
        \centering
        \includegraphics[width=\linewidth]{figs/vgg_taskonly.pdf}
        \caption{$L_{task}$}
        \label{fig:dista}
    \end{subfigure}
        \begin{subfigure}{.45\textwidth}
        \centering
        \includegraphics[width=\linewidth]{figs/vgg.pdf}
        \caption{$L_{task} + L_{mask}$}
        \label{fig:distc}
    \end{subfigure}
    \caption{VGG model with masking with different losses. Blue bars are the encoder part and red bars are the decoder part. Deeper features in the encoder are more prone to removal due to high redundancy. Decoder layers of the same number as the encoder are also more prone to removal which supports multiple other proposed engineered architectures.}
    \label{fig:dist}
\end{figure*}

\section{Masks analysis.} Figure \ref{fig:dist} shows the number of features removed per layer and the original number of filters for two pruned models optimizing only task loss or both task and sparsity loss. Figure \ref{fig:dista} shows VGG layers with masking optimized only with $L_{task}$ and interestingly the most removed features are in the deeper layers in the decoder. This aligns with PyDnet pyramidal design, where they predict the depth at different levels using shallow or lightweight decoders. The observation still holds as we apply $L_{sp}$ (Figure \ref{fig:distc}) on the final loss where layers with the same number of filters are removed more from the decoder part rather than the encoder part. This observation gives more insight on building encoder-decoder models where usually similar number of weights of the encoder part is used for the decoder which we argue this does not need to be the case. To the best of our knowledge, there is no previous work discussing the encoder-decoder choice of design.

\section{weights sparsity vs masks sparsity.} We evaluated our method compared to training with $\ell$1-norm regularizer on the convolutional filters weights. Not only sparsifying the weights will still require extra step based on a feature importance criteria (e.g remove all weights with norm $\leq$ threshold) for the weights to be pruned and then retraining, but also it does not have some favorable properties as in our method. $\ell$1-loss on all the weights do not differentiate between weights in early layers and those on later layers. However in our $L_{mask}$ from Eq. 2, the layers with larger number of filters contribute more to the loss. It makes sense to try to compress more in wide layers as filters tend to be more redundant than thin layers. Although, a downside to this property is as training progresses and maximum compression rate for a layer is reached, the loss does not adapt to the fact that these wide layers are thinner and might have less redundant features now. This only affect the maximum possible compression rate that can be reached rather than accuracy. As a sanity check we trained VGG with $\ell$1-regularization, we set regularization weight $\lambda$ = 0.005 and found the network having hard time to converge with D1 all error reaching 50.282 on KITTI split.

\bibliographystyle{IEEEbib}
\bibliography{strings,refs}